# Super-Efficient Spatially Adaptive Contrast Enhancement Algorithm for Superficial Vein Imaging

A. M. R. R. Bandara
University of Moratuwa, Katubedda,
Moratuwa, Sri Lanka.
ravimalb@uom.lk

K. A. S. H. Kulathilake
Rajarata University of Sri Lanka
Mihintale, Sri Lanka.
kulathilakeh@as.rjt.ac.lk

P. W. G. R. M. P. B. Giragama
Base Hospital Kebithigollawa
Kebithigollawa, Sri Lanka.
mgiracn@yahoo.com

# Super-Efficient Spatially Adaptive Contrast Enhancement Algorithm for Superficial Vein Imaging

A. M. R. R. Bandara
University of Moratuwa, Katubedda,
Moratuwa, Sri Lanka.
ravimalb@uom.lk

K. A. S. H. Kulathilake
Rajarata University of Sri Lanka
Mihintale, Sri Lanka.
kulathilakeh@as.rjt.ac.lk

P. W. G. R. M. P. B. Giragama
Base Hospital Kebithigollawa
Kebithigollawa, Sri Lanka.
mgiracn@yahoo.com

*Abstract*— This paper presents a super-efficient spatially adaptive contrast enhancement algorithm for enhancing infrared (IR) radiation based superficial vein images in real-time. The super-efficiency permits the algorithm to run in consumer-grade handheld devices, which ultimately reduces the cost of vein imaging equipment. The proposed method utilizes the response from the low-frequency range of the IR image signal to adjust the boundaries of the reference dynamic range in a linear contrast stretching process with a tunable contrast enhancement parameter, as opposed to traditional approaches which use costly adaptive histogram equalization based methods. The algorithm has been implemented and deployed in a consumer grade Android-based mobile device to evaluate the performance. The results revealed that the proposed algorithm can process IR images of veins in real-time on low-performance computers. It was compared with several well-performed traditional methods and the results revealed that the new algorithm stands out with several beneficial features, namely, the fastest processing, the ability to enhance the desired details, the excellent illumination normalization capability and the ability to enhance details where the traditional methods failed.

## I. INTRODUCTION

Superficial vein imaging becomes a mandatory requirement for numerous applications including phlebotomy, intravenous catheter insertion procedures, varicose vein treatments and vein pattern based biometric authentication systems [1], [2]. The light absorption property of hemoglobin allows infrared (IR) imaging techniques to reveal the superficial veins without using an invasive procedure [3]. However, the high cost of professional IR based superficial vein detectors limits the availability of such an instrument in needy places. If a highly available consumer device such as a smartphone or a low-performance computer can be used for this purpose with a support of a dedicated IR image acquisition device, the cost can be reduced by a significant fraction due to the whole processing part can be done within the host device [4]. However, the complexity and the less time efficiency of the image processing algorithms prevent implementing such a low-cost solution. Therefore, this research focused on finding an efficient algorithm which can be executed in a highly available and low-performance computer such as a smartphone, in real-time but without losing any details which are necessary for the application.

Infrared images are known to have lower contrast and also frequently found as a composition of both over and under exposed regions. Therefore the traditional contrast enhancement techniques on raw IR images produce an unsatisfactory outcome. A typical solution to overcome this problem is by using a variant of Adaptive Histogram Equalization (AHE) [5], [6]. In AHE, the Probability Density Function (PDF) is obtained based on the neighborhood of the pixel in which the contrast to be enhanced. However, calculating the PDF for each and every pixel is computationally expensive considering the demand of this application. Further, AHE generally amplifies the noise because it has no way to regulate the contrast boost of the algorithm [7]. Although this noise amplification can be solved by using an improved method called Contrast Limited Adaptive Histogram Equalization (CLAHE) [8], the high computational cost still persists. Several other AHE variants such as plateau AHE, double plateau AHE and improved double plateau AHE have been proposed particularly for enhancing IR images [9]. However, all of these methods inherit the high computational cost of traditional AHE which is not bearable by a low-performance general purpose computer in a real-time application. Reference [7] suggests a technique called Modified Histogram Based Contrast Enhancement using Homomorphic Filtering (HM-FIL) which particularly tested with medical images. However, the homomorphic filter requires calculating the log of every pixel, transforming to the frequency domain, high-pass filtering and transforming back to the spatial domain, which requires enormous time for processing a single frame [10].

The proposed algorithm utilizes the response of low-frequency range of the IR image signal to shift the dynamic range of the given pixel. The dynamic range can be adjusted using a single parameter to limit the amplification. The proposed approach in this study is named as Speeded-Up Adaptive Contrast Enhancement (SUACE). The performance of the implemented algorithm was evaluated by means of the computation time, the entropy and the visibility of desired details, comparing with the CLAHE and HM-FIL. The evaluation was done by considering the IR images of three important areas of human body namely dorsum of hand, wrist and cubital fossa due to the frequent usage in phlebotomy.

## II. SPEEDED-UP ADAPTIVE CONTRAST ENHANCEMENT

The objective of the proposed algorithm is to efficiently enhance the contrast of superficial vein images while adapting the contrast boosting according to the illumination variance over different regions of the image. The algorithm analyzes the illumination variances first and then selects the reference dynamic range to apply a linear contrast stretching process.

*A. Analysis of Illumination Variances*

The linear contrast enhancement with a fixed dynamic range is less effective when the illumination varies significantly over different regions of an image. Fig.1. shows how the illumination varies in a typical IR image of superficial veins. The illumination map in Fig.1 was calculated by using a Gaussian smooth filter with setting the σ to 15. The SUACE algorithm assumes that the contrast can be enhanced optimally by scaling the pixel values from a reference range in which the boundary values proportional to the corresponding response from the illumination map, to an adjustable higher dynamic range.

The frequency cut-off can be tuned with the standard deviation i.e. σ in the Gaussian filter. Section C describes a simple approach to finding a suitable value for σ in this application.

*B. Dynamic Range Selection*

The SUACE algorithm uses a fixed width of range for enhancing the contrast all over the image. However, the range is shifted according to the calculated illumination, i.e. the two boundary values of the range are varied from point to point. The selection of boundary values can be done by

$$a(x,y) = g(x,y) - \frac{d}{2}$$
$$b(x,y) = g(x,y) + \frac{d}{2} \quad (1)$$

where *a(x,y)* and *b(x,y)* are the lower and upper boundaries of the reference range for the scaling, respectively and *d* is the width of the reference dynamic range, which should be optimized for the application. The *g(x,y)* in (1) is the response of low-frequency range of the raw image which is calculated by convolving the original signal *I(x,y)* with the Gaussian kernel as

$$g(x,y) = I(x,y) * f(x,y)$$
$$f(x,y) = \frac{1}{2\pi\sigma^2} e^{-\frac{x^2+y^2}{2\sigma^2}} \quad (2)$$

The reference range of the dynamic range enhancement is shifted as it symmetrically lies around the illumination reference point i.e. allowing *d/2* space on both sides, in order to give a similar probability of the pixel intensities within the reference range, which significantly contribute to the desired details. However, the range can be still selected asymmetrically if the spatial dimension of the desired detail to be enhanced is much smaller than the filter size. The superficial veins appear dark relative to the surrounding but the average width of a vein is always around 50% of the size of selected Gaussian kernel hence the symmetric splitting of the range is used.

The dynamic range enhancement is done with linear contrast stretching using the calculated *a(x,y)* and *b(x,y)* contrast limit boundary points as shown in

$$\hat{I}(x,y) = \begin{cases} 0 & I(x,y) < a(x,y) \\ 1 & I(x,y) \geq b(x,y) \\ \frac{(I(x,y) - a(x,y))}{d} \times k, & \text{otherwise} \end{cases} \quad (3)$$

where *Î(x,y)* is the contrast enhanced image and *k* is the width of the new dynamic range.

*C. Parameter Selection*

The effective blur radius of the Gaussian kernel depends on the value of σ as shown in (2). The σ in SUACE algorithm controls the resolution of the illumination map. The resolution can be increased by selecting a smaller value for the σ which decreases the kernel radius and vice versa. The higher resolution of illumination map in SUACE tends to boost spatially smaller details because of the illumination response gets closer to the pixel value. Lower resolution illumination map can suppress small size details but preserve larger details. Therefore, the kernel size should be selected as it is larger than the size of desired detail which is to be amplified. It is learned that setting σ to 1/3 of the average width of desired detail performs fairly well. However, a fine-tuning might be needed for the optimal outcome.

*D. New Dynamic Range Selection*

The new dynamic range *d* has to be tuned according to the environment because it needs to enhance the desired details to a visible level without boosting the noise. However, the noise in this application is not affecting severely because the final result of this application is not used for a quantitative analysis. All the experiments in this work were carried out by setting the new dynamic range to 21.

### III. EXPERIMENTAL SETUP

The SUACE algorithm was implemented and installed in a commercial grade smartphone and tested using an IR camera. It was evaluated by comparing with CLAHE and HM-FIL, which were implemented on the same testing platform. This section describes the instrument and application configuration with details of the evaluation.

*A. IR Image Aquisition*

A VGA resolution CMOS sensor with a 2.8mm lens is used in the prototype for the image acquisition. The camera is connected with USB video class (UVC) compliant driver. An IR band-pass filter (808nm-1064nm) is used in front of the lens to enhance the contrast of the image. A set of IR LEDs with a diffuser sheet is used to light up the subject.

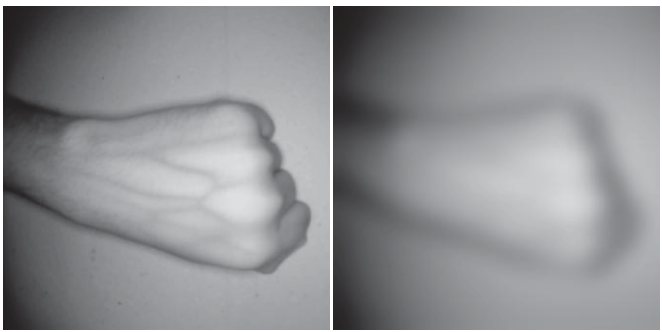

Fig. 1. The IR raw image in the left and its illumination map where σ=15 in the right.

*B. Host Device Configuration*

The whole processing part is done inside a smartphone powered by an Android operating system. The device is equipped with 4x2.1 GHz Cortex-A57 and 4x1.5 GHz Cortex-A53 CPUs with approximately 4GB of RAM and Mali-T760MP8 GPU. The IR camera is connected to the host device through USB. The video is streamed into the application using the video for Linux (V4L2) driver which is an inbuilt driver in the Android operating system for accessing UVC compliant devices. Fig. 2 shows the arrangement of the apparatus of the prototype.

The image processing modules of the application were built to native codes and linked with the Android application to facilitate the user interface. The native implementation allows the real-time streaming of video in an idle state at the rate of 30 frames per seconds when no additional processing task is executed.

The three methods such that the SUACE and other two comparable approaches were implemented in three separate applications, each executes alone at a time.

*C. Evaluation*

The SUACE algorithm is evaluated by means of the time efficiency, the clarity of the visual appearance of superficial veins and overall visual quality compared with the other two comparable approaches. Besides these facts, the effect of different dynamic ranges and the resolution of illumination map also evaluated. The time efficiency is measured by the number of frames which can be processed per seconds relatively to the idle frame rate as a percentage. This measurement is easily comparable and directly quantifies how fast the process in practice. The entropy value is used to compare the contrast enhancement quantitatively [11]. Mathematical expression to calculate the entropy of an image is given in

$$E = -\sum_{n=0}^{N-1} P(n) \log_2 P(n) \qquad (4)$$

where *P(k)* is Probability Mass Function (PMF) of the histogram of the resultant image. The clarity of superficial veins is evaluated by comparing the regions where weakest details are found in the resultant frames of the 3 approaches.

The experiment was done by capturing images of 6 males and 6 females who are having significantly different skin colors, vein patterns, hair density and skin fat levels. All participants were in normal hydration level and normal blood pressure. The experiment was repeated for three important areas of the human body i.e. dorsum of hand, wrist and cubital fossa with its surroundings. The set of images presented in this paper was selected from a male and female participant who is having a significantly different visibility of veins in the raw image.

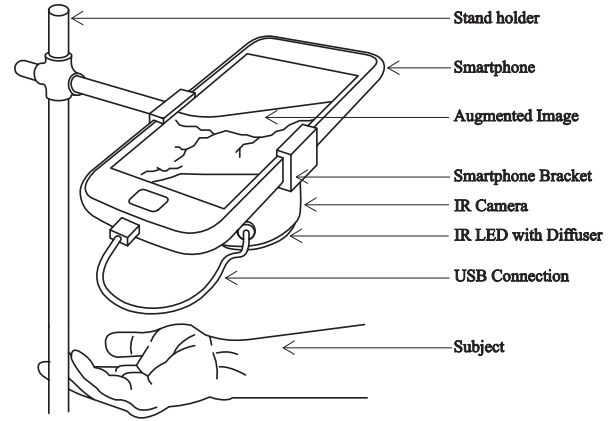

Fig. 2. The arrangement of the apparatus in the prototype device.

The experiment was carried out by placing the subject a 15cm away from the camera. A stage with IR diffusible surface was used as the background of the subject to enhance the clarity. The complete set of apparatus was set up in an indoor environment lit by a 50W compact fluorescent light.

IV. RESULTS AND DISCUSSION

The result of the experiment carried out to tune the parameters of SUACE algorithm is presented in Fig. 3, and the performance of the algorithm with the best parameters is presented in Fig. 4.

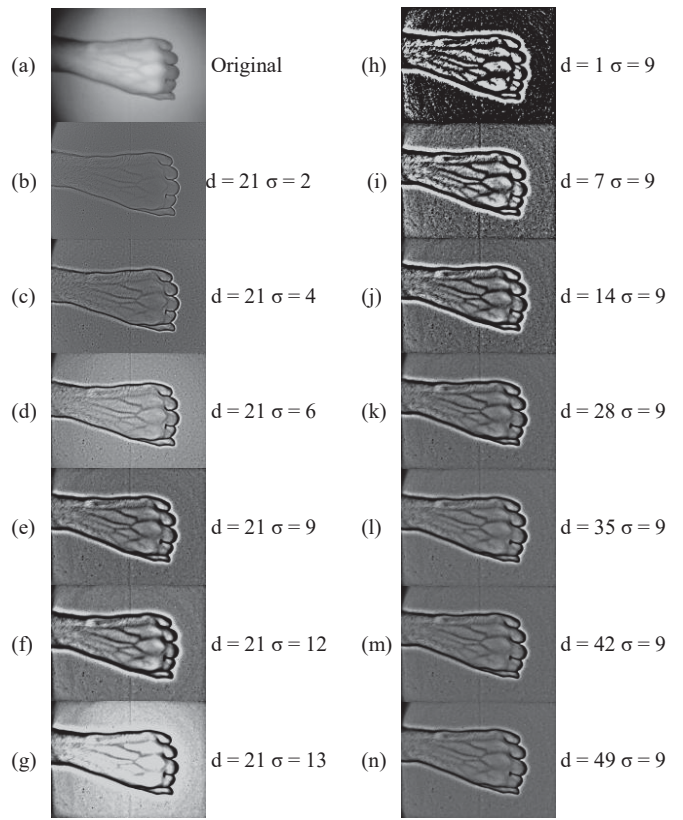

Fig. 3. Effect of different combinations of *d* and *σ*. (a) Original image; (b)-(g) Resultant images while varying σ keeping *d* static. (h)-(n) Resultant images while varying *d* keeping σ static by setting to 9.

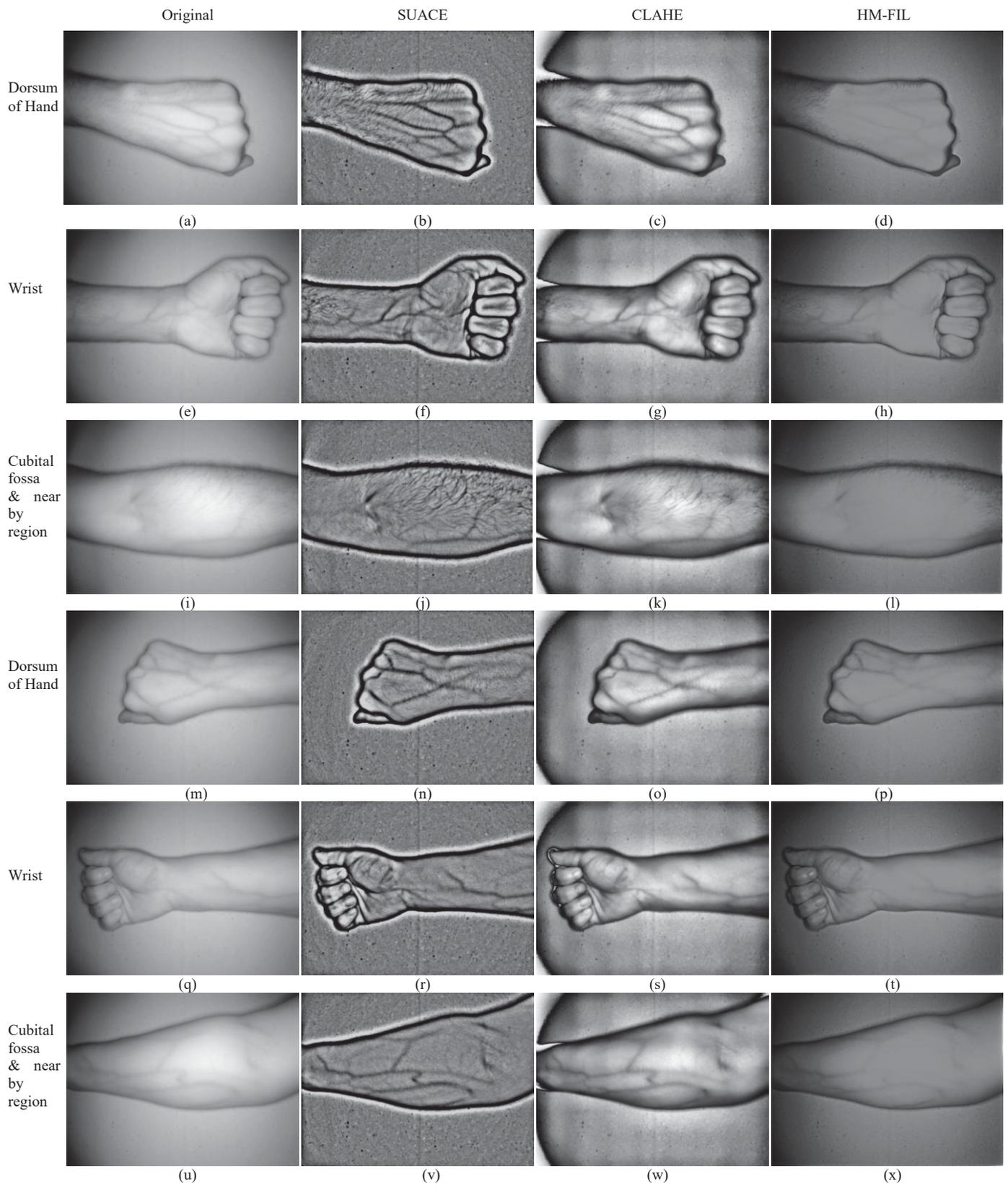

Fig. 4. Enhanced images of three selected body regions of a male and female participant using the three approaches. (a) - (l) The images of male participant. (m) - (x) The images of the female participant.

## A. Results of Parameter Tuning Experiment

Fig. 3 shows the outputs for a different combination of parameters. The result clearly shows that the effect of *d* and *σ* to the output is mutually independent. Several instances of the *d* and *σ* showed reasonable results, therefore one of the best pairs of parameters, i.e. *d*=21 and *σ*=9 were picked to use in the remaining experiments.

## B. Image Quality Assessment

Table I shows results of entropy values on a sample set of given images, processed by different methods. In general, the higher the entropy is, the richer details and information the image holds. When the contrast is enhanced, only the desired details should be boosted and all the other details will be suppressed helping the required details to stand out.

Therefore better the contrast enhancement; the entropy should be reduced even though the reverse is not true always. As the result shows the average entropy is least for the SUACE algorithm where CLAHE shows the highest. The significantly lower entropy expresses the outstanding detail suppression power of the proposed algorithm. However, the entropy along does not express how better the desired details were enhanced. Therefore a qualitative analysis was done.

Fig. 4 shows the contrast enhancement results of the three approaches with several sample images. The α in HM-FIL [7] had been set to 0.95 which provided a good illumination balance. This well-balanced illumination distribution can be clearly seen in the results as well. However, the visual result clearly shows that HM-FIL is not capable of enhancing the veins although it showed better entropy compared to CLAHE. The reason for the better entropy might be the suppressed information due to the illumination balancing, but it has to be verified by another experiment.

TABLE I.  ENTROPY VALUES OF THE ORIGINAL AND THE RESULTANT IMAGES OF THE THREE CONTRAST ENHANCEMENT METHODS

| Image Name | Original | SUACE | CLAHE | HM-FIL |
|---|---|---|---|---|
| P1_dorsum_hand | 7.2 | 3.4 | 7.6 | 6.5 |
| P1_wrist | 7.1 | 3.5 | 7.5 | 6.4 |
| P1_cubital_fossa | 7.2 | 3.3 | 7.6 | 6.6 |
| P2_dorsum_hand | 7.2 | 3.4 | 7.6 | 6.5 |
| P2_wrist | 7.1 | 3.4 | 7.5 | 6.5 |
| P2_cubital_fossa | 7.2 | 3.5 | 7.4 | 6.6 |
| P3_dorsum_hand | 7.2 | 3.3 | 7.5 | 6.5 |
| P3_wrist | 7.1 | 3.4 | 7.6 | 6.5 |
| P3_cubital_fossa | 7.3 | 3.4 | 7.5 | 6.4 |
| P4_dorsum_hand | 7.2 | 3.3 | 7.5 | 6.3 |
| P4_wrist | 7.1 | 3.4 | 7.6 | 6.5 |
| P4_cubital_fossa | 7.3 | 3.4 | 7.5 | 6.6 |
| **Average** | **7.18** | **3.39** | **7.53** | **6.49** |

Although the entropy does not verify the true potential of CLAHE, it still performs better than HM-FIL with enhancing veins. When the results of CLAHE are compared with the results from SUACE, both are acceptable up to a certain level. However, considering the clarity of majority of the veins, the results of SUACE will be more preferable. The images of the dorsum of hand always show that the majority of veins are enhanced similarly in the both approaches, but when the cubital fossa is considered, CLAHE seems missed several veins partially and a few in completely where SUACE had boosted them all relatively better. The sub figures (j) and (k) in Fig. 4 proves the fact that SUACE has accurately enhanced the veins within the cubital fossa yet the same set of veins could not be identified in the result of CLAHE. When (f) and (g) in Fig. 4 are compared, CLAHE could only enhance veins at the region near the palm, whereas SUACE had enhanced the veins all over the visible part of the forearm. All the images captured from the 12 participants exhibited this superior performance of SUACE but only the results of two participants were included in this paper due to the space limitation.

Another significant property which can be seen in all the images enhanced by SUACE is that the uniform illumination distribution all over the image. CLAHE has amplified the boundary of the shadowed region and the well-illuminated region whereas the SUACE has safely removed that boundary. Moreover, SUACE still has enhanced the vein like details in all three regions i.e. the shadowed region, well-illuminated region and the region which covers their boundary.

When the experiment was in the process, the frames of the video, which were processed by CLAHE showed a strange artifact which is a square shaped pattern of illuminations. This pattern had been identified as one of the reasons for the inconsistency in contrast distribution over different regions of veins. The pattern persisted statically and when a detail i.e. a vein comes into a boundary of the repeating units in the pattern, the contrast is lowered on the corresponding location and the details will be faded out. Note that this effect cannot be seen in static images including the samples given in Fig. 4 but clearly visible with disturbing the vision of veins in the video. Fig. 5 clearly shows this artifact which became a major drawback of CLAHE to use in this application.

Fig. 4 shows that SUACE significantly enhanced the hair in addition to the network of veins, compared to the results of CLAHE. This can be considered as a limitation of SUACE algorithm. However, the veins can be distinguished from hair by applying a slight pressure to the indistinguishable vein to temporarily stop the blood flow because the bloodless veins are not visible whereas the changes in the blood flow do not impact on the visibility of hair.

## C. Computation Time Assessment

Table II shows the average frames per seconds which can be processed with each of the algorithms relatively to the idle frame rate, as a percentage. SUACE achieved the best computation time compared to the other two algorithms. The histogram modification, histogram equalization and the usage of homomorphic filter justify the worst computation time of HM-FIL. Although CLAHE showed a better computation time compared to HM-FIL, the speed of computation of SUACE is still significant when the targeted application is taken into the consideration because how close to the real-time processing means the better accuracy of the medical procedure.

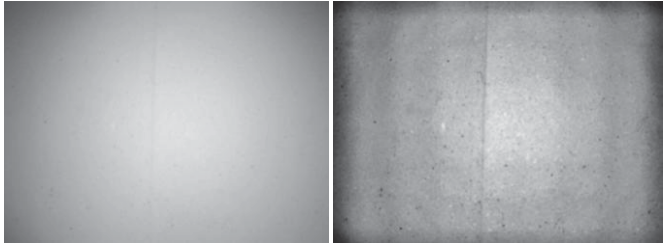

Fig. 5. Original image left and the square shaped artifact of its CLAHE Enhancement in right.

TABLE II. AVERAGE FPS WHILE NO PROCESS AND THE THREE CONTRAST ENHANCEMENT PROCESSES APPLIED SEPARATELY

| Algorithm | Average FPS | Relative FPS |
|---|---|---|
| Idle (No Process) | 29.8 | 100% |
| SUACE | 28.4 | 95.3% |
| CLAHE | 25.5 | 85.6% |
| HM-FIL | 18.2 | 61% |

## V. CONCLUSION

This paper presented a novel approach of super-efficient adaptive contrast enhancement technique for real-time superficial vein imaging. The approach targets the implementation in highly available low-performance computer such as a smart phone. The experimental results obtained using the android based prototype equipment, proved the excellence of the proposed algorithm by means of the entropy analysis, the clarity of superficial veins, illumination normalization and the computation efficiency. Future works can be directed to find other potential applications of SUACE.